\RequirePackage{fix-cm}
\documentclass[twocolumn]{svjour3}          
\smartqed  
\usepackage{graphicx}
\usepackage[algo2e]{algorithm2e} 
\usepackage{algorithmic,algorithm}
\usepackage{hyperref}
\usepackage{amsmath,amssymb,amsfonts}
\usepackage{xcolor}
\DeclareMathOperator{\E}{\mathbb{E}}
\usepackage{hyperref}
\DeclareMathOperator{\atantwo}{atan2}

\begin{document}

\title{Generating Quality Grasp Rectangle using Pix2Pix GAN for Intelligent Robot Grasping
}

\author{Vandana Kushwaha\and Priya Shukla\and G C Nandi}

\institute{Vandana Kushwaha \at
              Center of Intelligent Robotics, Indian Institute of Information Technology Allahabad, Prayagraj-211015, U.P., INDIA \\
              \email{kush.vandu@gmail.com}  
          \and
          Priya Shukla \at
              Center of Intelligent Robotics, Indian Institute of Information Technology Allahabad, Prayagraj-211015, U.P., INDIA \\
              \email{priyashuklalko@gmail.com} 
          \and
          G. C. Nandi \at
              Center of Intelligent Robotics, Indian Institute of Information Technology Allahabad, Prayagraj-211015, U.P., INDIA \\
              \email{gcnandi@iiita.ac.in}
}

\date{Received: date / Accepted: date}

\maketitle

\begin{abstract}
Intelligent robot grasping is a very challenging task due to its inherent complexity and non availability of sufficient labelled  data. Since making suitable labelled  data available for effective training for any deep learning based model including deep reinforcement learning is so crucial for successful grasp learning, in this paper we propose to solve the problem of generating grasping Poses/Rectangles using a Pix2Pix Generative Adversarial Network (Pix2Pix GAN), which takes an image of an object as input and produces the grasping rectangle tagged with the object as output. Here, we have proposed an end-to-end grasping rectangle generating methodology and embedding it to an appropriate place of an object to be grasped. We have developed two modules to obtain an optimal grasping rectangle. With the help of the first module, the pose (position and orientation) of the generated grasping rectangle is extracted from the output of Pix2Pix GAN,  and then the extracted grasp pose is translated to the centroid of the object, since here we hypothesize that like the human way of grasping of regular shaped objects, the center of mass/centroids are the best places for stable grasping. For other irregular shaped objects, we allow the generated  grasping rectangles as it is to be fed to the robot for grasp execution. The accuracy has significantly improved for generating the grasping rectangle with limited number of Cornell Grasping Dataset augmented by our proposed approach to the extent of  87.79\%. Rigorous experiments with the Anukul/Baxter robot, which has 7 degrees of freedom (DOF), causing redundancy have been performed. At the grasp execution level, we propose to solve the inverse kinematics problems for such robots using Numerical Inverse-Pose solution together with Resolve-Rate control which proves to be more computationally efficient due to the common sharing of the Jacobian matrix. Experiments show that our proposed generative model based approach gives the promising results in terms of executing successful grasps for seen as well as unseen objects (\href{https://www.youtube.com/watch?v=2C9QBbBSxUk}{refer to demonstrative video}).

\end{abstract}

\section{Introduction}
\label{intro}
Being human, grasping is a very easy and intuitive action for us as we have the ability to learn using our past experiences. A child, for example, has poor grasping skill compared to an adult who learns over the years how to grasp object more skillfully. However, imparting the same experience based learning in a robot is a very challenging problem to solve. Although a lot of work has been done in the area of robotic grasping \cite{jiang2011efficient,kumra2017robotic,lenz2015deep,Nandi2018DeepLB,redmon2015real,saxena2008robotic,saxena2010vision}, robots are still far from human level performances, specially for grasping unseen objects. The robotic grasping task depends on the three important and sequential phases. The very first and primary phase is the detection of potential grasp which requires an understanding of the object to be grasped from a given scene. The 3-dimensional vision sensor system helps the robot to get the information about its environment. From obtained sensor information, predicting the pose for potential grasp and mapping it to the real-world coordinates is the most critical step because other phases depend on this phase. In the second phase, an optimal trajectory is planned using calculated real-world coordinates for the robotic arm to reach the predicted grasp pose. Then, planned trajectory is executed by the robotic arm to grasp an object in the last phase.

\par

This paper mainly focuses on the problem of predicting optimal grasping rectangles for seen as well as unseen objects over a given RGB image using the generative model. We are using Pix2Pix Generative Adversarial Network (Pix2Pix GAN)
\cite{pix2pix2017} for generating the grasping rectangle directly, instead of sampling and then finding the best one from the sampled candidate rectangles. 
Here, Pix2Pix GAN takes image of an object as an input and produces the grasping rectangle tagged with the object as output. In this research, we have  developed two modules: first is the ``Pose Extraction" module, which is used for extracting the position and orientation of the generated rectangle from the output image of Pix2Pix GAN and second is the ``Pose Translation" module, which is used for translating the extracted grasp pose to obtain an optimal grasping rectangle. For a given object, the chances of successfully grasping has increased  with the help of an optimal grasping rectangle. Our model has achieved  the accuracy of 87.79\% for predicting the grasping rectangle correctly on the Cornell Grasping Dataset  \cite{cgd}. \par

Major contributions of our work are:
\begin{itemize}
    \item We present a generative model based approach for generating the grasping rectangle for seen as well as unseen objects. To the best of our knowledge, this is the first time Pix2Pix GAN has been applied in the domain of robotic grasp generation.
    
    \item We have developed two modules named as: Pose Extraction and Pose Translation. With the help of developed modules,  the pose of the grasping rectangle generated by Pix2Pix GAN is extracted and then, the extracted grasp pose is translated to obtain an optimal grasping rectangle.
    
    \item Our work can be used to create an annotated dataset for robotic grasping  using RGB images only.

    \item For evaluating the performance of our proposed approach, grasping experiment is performed with the Baxter research robot (named as Anukul) which has 7 degrees of freedom (DOF), causing redundancy. We propose to solve this problem using Numerical Inverse-Pose solution together with Resolve-Rate control which proves to be more computationally efficient due to the common sharing of the Jacobian matrix. 
\end{itemize}

Other sections of the paper are formulated as follow: 
Section 2 discusses the related work proposed so far in the area of robotic grasping, Section 3 explains the problem formulation, Section 4 covers the preliminaries on GAN and Pix2Pix GAN which are important ingredients for our proposed strategy, Section 5 discusses the experimental setup along with the results and  its analysis for our proposed approach, and lastly the Section 6 presents the conclusion and future work.

\section{{Related Work}}
A plenitude of different techniques has been explored for decades to solve the problem of robotic grasping. These techniques can be primarily categorized into two approaches: analytical planning based approach and empirical learning based approach \cite{bohg2013data,sahbani2012overview}. The analytical approaches, being planning based, determine the grasp using complex models of geometry, dynamics, and kinematics \cite{bicchi2000robotic,shimoga1996robot}. Due to the intrinsic complexity of the planning based analytical approach, it is very hard to establish the physical interactions between a manipulator and an object in the real world unstructured environment.
\par 
On the contrary, the empirical learning based approaches are based on the estimation and the experience-based model for predicting the grasp. In finding good grasp points, certain techniques work well  for familiar objects as shown in \cite{detry2009autonomous,el2008handling,goldfeder2007grasp}. However, they are not suitable for the unseen (novel) objects.\par

Grasp prediction for novel object has achieved a major improvement with the augmentation of vision-based deep learning techniques and empirical based approaches   \cite{lenz2015deep,mahler2017dex,Nandi2018DeepLB,pinto2016supersizing,wang2016robot}. Most of these techniques use a two-stage pipeline: firstly, from input image or point cloud, generate grasp candidates and then obtain the best one based on its ranking given by Convolutional Neural Network (CNN). Once the best grasp is finalized, the robot executes it in an open-loop manner without taking any feedback. It requires the precise control and calibration between static environment, robot, and camera.
Another approach in which the sliding window concept is used with large neural networks for grasp synthesis. In such  frameworks, a number of a small patch of the input image is fed to the classifier to predict the potential of grasp for an object. The classifier (CNN) assigns a score to patches that decide its potential for a good grasp. The major drawback of these approaches is the computational time which is inappropriate for a real-time system. \par
In contrast to above approaches,  \cite{kumra2017robotic} and \cite{redmon2015real}  uses the Deep CNN (AlexNet and ResNet) to regress the best grasp pose directly for an input image. However, single best grasp produced by regression for an entire image may produce the average of all possible grasps, which might not be a good potential grasp of an object.
In \cite{morrison2018closing}, Morrision et al. proposed a model known as Generative Grasp Convolution Neural Network (GG-CNN) to address the issues of execution time and averaging of all possible grasp. For every pixel of an image,  GG-CNN predict the quality of grasp and then a high quality grasp pose is selected for the execution in closed loop manner. In \cite{mahajan2020semi}, a semi-supervised approach is used for learning to grasp.
Training of vision-based model for grasp detection is a very tedious task as it has insufficient amounts of labelled  data. Therefore,in the present investigation we have proposed a new approach to predict the grasping rectangle for an object using generative model, which takes images as input and output the grasping rectangles tagged with the input images. Our model can generate data for the robotic grasping and solve the problem of limited training dataset. To validate the learning through successful grasp executions, we have also performed rigorous experiments using Anukul robot (a 7-DOF robot having redundancy), using Numerical Inverse-Pose solution together with Resolve-Rate control which proves to be more computationally efficient and hence more suitable for real time applications due to the common sharing of the manipulator Jacobian matrix.

\section{Problem Formulation}

\begin{figure}[!ht]
 \centering
    \includegraphics[scale =0.4]{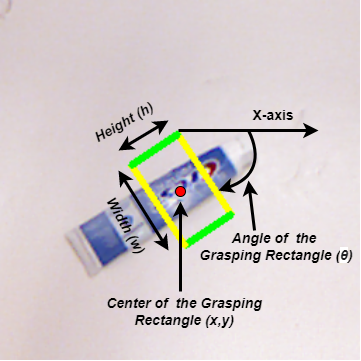}
    \caption{A 5D Grasp Pose representation using position and orientation. The green lines show the gripper's orientation, while yellow lines show maximum distance between gripper plates before the grasp execution. }
    \label{fig:1}
\end{figure}

In this research, we are focusing on the problem of predicting an optimal grasp pose to perform grasping from the given RGB image of an isolated object. A grasp pose $g$ is defined by (\ref{eq:1}), 
where $ (x,y)$ is the center of the grasp rectangle, $\theta $ is an orientation of the grasp rectangle from X-axis, $(h,w)$ is the height and the width of the grasp rectangle as shown in Fig. \ref{fig:1}.
\begin{equation}
    g = (x,y,\theta,h,w)
    \label{eq:1}
\end{equation}
\par
For given RBG image I, $g_{I}$ is a grasp pose in an image space and $g_{R}$ is the grasp pose of the object in the robot configuration space.
With the help of transformation functions, $g_{R}$ can be obtained from $g_{I}$ using (\ref{eq:2}), where ${}^{C}T_{I}$ transform 2D image coordinates to 3D camera coordinates and  $ {}^{R}T_{C}$ transforms from camera  configuration space to robot  configuration space.
\begin{equation}
    g_{R}= \Big[ {}^{R}T_{C} \big[ {}^{C}T_{I}(g_{I}) \big] \Big]
    \label{eq:2}
\end{equation}

\section{Methodology}

\subsection{Preliminaries}

\subsubsection{Generative Adversarial Network (GAN)}
\begin{figure}[!ht]
\centering
    \includegraphics[scale =0.4]{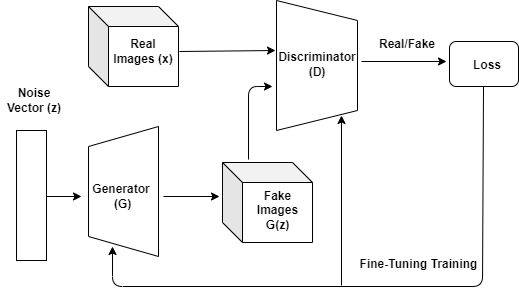}
    \caption{The architecture of GAN}
    \label{fig:2}
\end{figure}

It is one of the most popular frameworks of generative models proposed by Ian Goodfellow et al. in 2014  \cite{goodfellow2014generative}. It consists of two deep neural networks: generator, which tries to generate data as real as possible, and discriminator, which identifies the data as a real or a fake as shown in Fig. \ref{fig:2}. Both the models compete and co-operate with each other during training using zero sum game theory,  to reach their optimal performance. Vanilla GAN, although is plagued with many problems such as non convergence, mode collapse and vanishing gradient, has produced outstanding results for generating data like text, image and video.
Other variant of GAN has also been investigated for real-life application like image-to-image translation, text-to-image synthesis etc. 

The discriminator model of GAN tries to maximize the probability of classifying data correctly whether it belongs to the real dataset or data generated by the generator. The objective function for discriminator has been shown in (\ref{eq:3}), where D(x) denotes the output of discriminator.
\begin{multline}
  \max_{D} V(D)= \E_{x \sim P_{data}(x)} \Big[ \log D(x) \Big]\\ +  \E_{z \sim P_z(z)} \Big[ \log \Big( 1-D \big( G(z) \big) \Big) \Big]
  \label{eq:3}
\end{multline}
First term in (\ref{eq:3}) helps the discriminator to recognize real images better and second term helps to recognize generated images better.\par
In contrast, the generator tries to fool the discriminator by minimizing the probability of identifying the data generated by the generator correctly.  Let G(z) denotes the output of the generator and its objective function looks like (\ref{eq:4}).
\begin{equation}
  \min_{G} V(G)=  \E_{z \sim P_z(z)}\Big[ \log \Big( 1-D \big( G(z) \big) \Big) \Big]
  \label{eq:4}
\end{equation}
On combining the  objective functions of both the models, GAN objective function turns out to be a  min-max game between discriminator D and generator G in which the discriminator tries to maximize and the generator tries to minimize the loss function for GAN as shown in (\ref{eq:5}).
\begin{multline}
 \min_{G} \max_{D} V(D,G)= \E_{x \sim P_{data}(x)} \Big[ \log D(x) \Big] \\ + \E_{z \sim P_z(z)} \Big[ \log \Big( 1-D \big( G(z) \big) \Big) \Big]
 \label{eq:5}
\end{multline}

During training, both the models learn collectively by alternating gradient descent. Firstly, the discriminator model perform one iteration of gradient descent using real images and generated images with fixed parameters of the generator model. Subsequently, it flips the side and train the generator model for another one iteration with fixed parameters of the discriminator model. This process continue till the generator  model learns to produce quality images.\par
Some major advantages of GAN are:
\begin{itemize}
    \item  GAN is capable of learning internal representation of data. So with the help of unlabelled  data GAN can be trained which is a great advantage since capturing labelled  data for robot grasping which is a manual process and requires a lot of time.
    \item Real power of GAN is the ability to generate the different kinds of data like images, text, audio, and video which is identical to real data.
    \item GAN can learn the internal representation of data without considering the data distribution i.e. messy or complicated.
\end{itemize}

In-spite of having many advantages, there exist some problems with the basic GAN architecture such as non-convergence of models, mode-collapse, vanishing gradient problem, and problem of  finding the Nash Equilibrium.
Consequently, to mitigate some of these problems, GAN with different cost functions have been tried as  shown in  \cite{arjovsky2017wasserstein,gulrajani2017improved,mao2017least,metz2016unrolled,mirza2014conditional}. Some new techniques have been  proposed in \cite{arjovsky2017towards,salimans2016improved} and some changes are made in the architecture \cite{radford2015unsupervised} to make it more effective and stable.

\subsubsection{Pix2Pix GAN} 

Pix2Pix GAN is based on the concept of conditional GAN (cGAN) \cite{mirza2014conditional}. It is used for image-to-image translation, in which a given image is transformed from one domain to another domain in a controlled manner.
For a given input image, the generator model of Pix2Pix GAN produces a translated version of the image as an output, while the discriminator model of Pix2Pix GAN takes real or generated paired images as input and determines the given pair as real or fake. \par

\begin{figure}[!ht]
    \centering
    \includegraphics[scale =0.22]{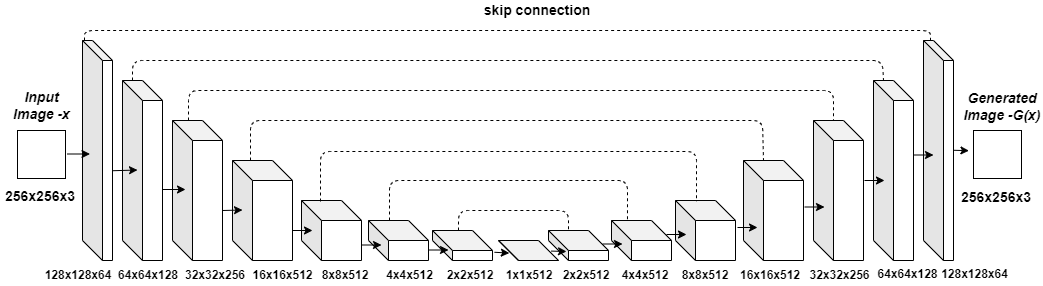}
    \caption{The architecture of U-Net Generator}
    \label{fig:3}
\end{figure}
The generator model, which consists of an Encoder-Decoder architecture, takes images as input instead of random noise and performs down-sampling of input images till bottleneck layer, and then representation is up-sampled to few layer to obtain desired sized output images. Instead of common encoder-decoder architecture, Pix2Pix GAN uses U-Net architecture for the generator model which is slightly different from encoder-decoder architecture, as it uses skip-connection between the layers of same size to avoid the bottleneck as shown in Fig. \ref{fig:3}.

In Pix2Pix GAN, the discriminator model is implemented as a PatchGAN. A PatchGAN is constructed as a deep convolution neural network to perform the conditional-image classification. It takes paired image (source images and target images) as input and classify the target image as real of fake translation of the source image, whereas a normal discriminator model classifies an entire input image as real or fake.

\begin{figure}[!ht]
    \centering
    \includegraphics[scale =0.21]{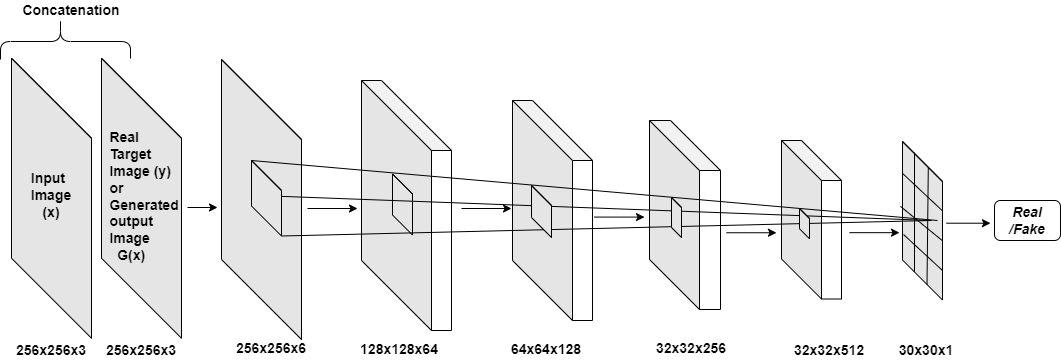}
    \caption{The architecture of PatchGAN Discriminator}
    \label{fig:4}
\end{figure}

 Further, a PatchGAN network, which is based on the concept of effective receptive field, predicts for each $N \times N$ patch from an input image as real or fake and then it averages out all the responses to produce a final output. The effective receptive field defines the relationship between patch  on the input image and one output activation of the model. The architecture of the PatchGAN discriminator model is shown in  Fig. \ref{fig:4} .

\begin{figure}[!ht]
\centering
    \includegraphics[scale =0.33]{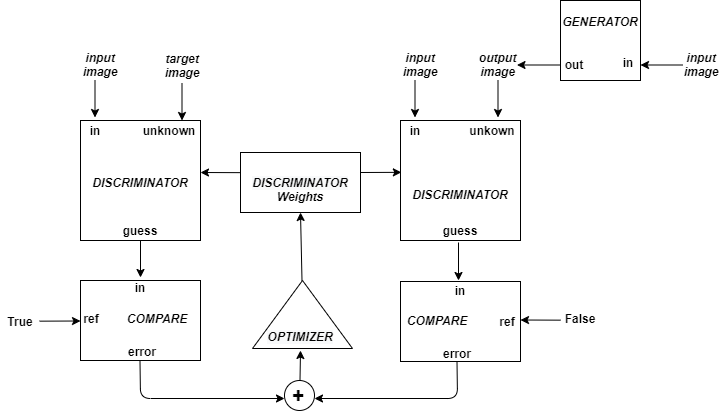}
    \caption{Training of the PatchGAN Discriminator}
    \label{fig:5}
\end{figure}

\begin{figure}[!ht]
 \centering
    \includegraphics[scale =0.38]{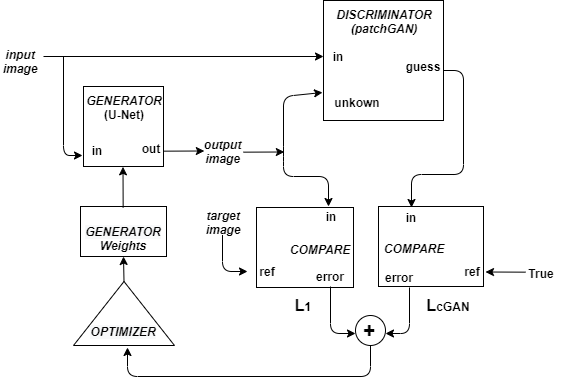}
    \caption{Training of the U-Net Generator}
    \label{fig:6}
\end{figure}

The training of the discriminator model of a PatchGAN is very much similar to the traditional GAN, which tries to minimize the error in prediction of the input image as real or fake. Fig. \ref{fig:5} shows the training process of the PatchGAN  discriminator and how the calculated error is back-propagated to minimize it. The Discriminator Loss (DL) is halved by using (\ref{eq:6}) to slow down the training process as it is very fast compared to the generator training.

 \begin{equation}
     DL= 0.5 * DL
     \label{eq:6}
 \end{equation}
For the training of the generator model, both the losses, $L_{cGAN}$ and $L_1$,  are used as shown in Fig. \ref{fig:6}. $L_{cGAN}$ is a adversarial loss of cGAN which  is defined by  (\ref{eq:8}) and $L_1$ is a mean absolute pixel difference between the expected target image and translated image of input image produced by the generator which has been defined by (\ref{eq:9}). Finally, the Generator Loss (GL) looks like (\ref{eq:7}), where, $\lambda$ is a hyper parameter which decides the impact of $L_1$ loss.
 \begin{equation}
     GL= L_{cGAN}+ \lambda{L1(G)}
     \label{eq:7}
 \end{equation}
 where, 
\begin{multline}
  L_{cGAN}(D,G)= \E_{x,y} \Big[ \log D(x,y) \Big] \\ + \E_{x,z} \Big[ \log \Big( 1-D \big( x, G (z,x) \big) \Big) \Big]
 \label{eq:8}
 \end{multline}
 and 
 \begin{equation}
 L_{1}(G)= \E_{x,y,z} \Big[ ||y- \big( G(z,x) \big)|| \Big]
 \label{eq:9}
\end{equation}

In comparison to other GANs, the benefit of Pix2Pix GAN is that it is relatively simple to implement  and it produces high-quality images with a large variety of image translation tasks. The generated images by Pix2Pix GAN seems to be clearer because of the use of both the losses: $L_1$ loss and adversarial loss of $L_{cGAN}$. 
\subsection{Proposed Methodology}

With the image-to-image translation skill of Pix2Pix GAN, a new approach is proposed to generate the grasping rectangles. Pix2Pix GAN takes an image of an object as input, and transformed it to an image of an object tagged with the grasping rectangle.  The whole process is divided into three major steps. Firstly, a paired image dataset is created using Cornell Grasping Dataset. Then, the pre-processing of created paired image dataset and training of  Pix2Pix GAN is performed.  Finally, the position and orientation of the generated grasping rectangle is extracted from output image of Pix2Pix GAN and then translated it to the centroid of the object to find an optimal grasp pose. After finding an optimal grasp pose in an image configuration space $g_{I}$, the grasp pose $g_{R}$, in robot configuration space is computed to execute the robotic grasp.

\subsubsection{Creating paired-image dataset from Cornell Grasping Dataset}
The paired-image dataset consists of an image from one domain along with its corresponding image from another domain. In our case, the image of an object is considered as one domain, and an image of an object tagged with the grasping rectangle is considered as another domain.

\begin{figure}[!ht]
\centering
    \includegraphics[width = 7 cm]{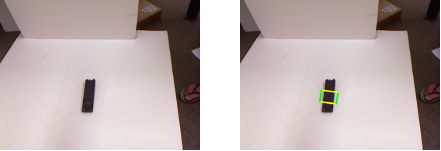}
    \caption{Hand labelled  positive grasp rectangle is drawn on an object for creating the paired-image dataset. The gripper plates orientation is shown by green sides of the grasping rectangle.}
    \label{fig:7}
\end{figure}

With Cornell Grasping Dataset, paired image dataset is created for the training of Pix2Pix GAN.  Cornell Grasping Dataset consists of 885 RGB images of 280 distinct objects with different orientations. Each image is associated with hand-labelled  positive rectangle text file, hand-labelled  negative rectangle text file, and point-cloud file.
In our approach image file along with its associated positive rectangle text  files has been used for creating the paired image dataset.

Using positive rectangle text files, the grasping rectangle is drawn on an image of an object. The sides of the grasping rectangle has yellow and green colors. The green sides of the rectangle represents the orientation of the gripper plates as shown in Fig. \ref{fig:7}. Cornell Grasping Dataset has 5,110 positive grasping rectangles, and hence 5,110 paired images are created for training the model.

\subsubsection{Training of Pix2Pix GAN for generating grasping rectangle}
 For the training of Pix2Pix GAN, paired image dataset has been used. Firstly, paired images are cropped to $256\times 256$  sizes for  feeding into the Pix2Pix GAN model as input. Then, we have trained Pix2Pix GAN with three different discriminators with the following patch sizes : $1\times 1$ patch discriminator, $16\times 16$ patch discriminator, and $70\times 70$ patch discriminator. 
\begin{figure}[!ht]
\centering
    \includegraphics[scale=0.4]{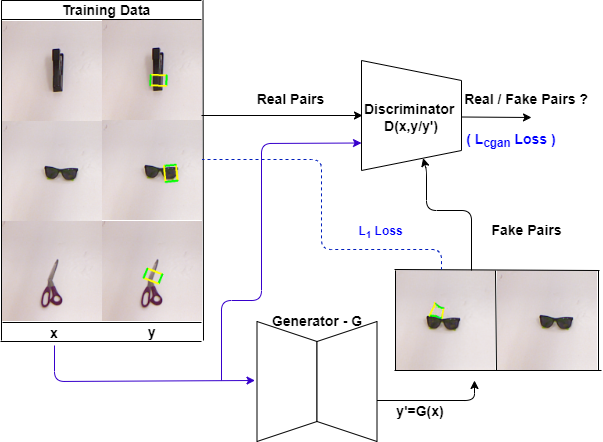}
    \caption{Training of Pix2Pix GAN}
    \label{fig:8}
\end{figure}

Fig. \ref{fig:8} shows the training process of Pix2Pix GAN with the paired image dataset. We have trained our model for $100K$ number of steps and saved the generator model after $1K$ steps. So, total 100 generator models are obtained after the training processes are completed for each patch size discriminator.
We have observed that among all the patch discriminators, Pix2Pix GAN with $70\times 70$ patch discriminator provide the best result. 

So, Pix2Pix GAN with $70\times 70$ patch discriminator is used to generated the grasping rectangle which  is translated to the centroid of the object for finding an optimal grasping rectangle which has been  discussed in the subsequent steps.

\subsubsection{Translation of Grasping rectangle Generated by Pix2Pix GAN to its optimal position}

For fine manipulations on the generated objects, translation of the generated grasping rectangle to its optimal position is required so as to increase the chances of successful grasp. To accomplish this task we have developed two modules namely, ``Pose Extraction" and ``Pose Translation". 
Firstly, the orientation and position of  the grasping rectangle generated by Pix2Pix GAN (with $70\times 70$ patch discriminator)  is extracted with the help of pose extraction module and then  extracted pose is translated  to the centroid of the object to obtained an optimal grasping rectangle using pose translation module. 

\begin{figure}[!ht]
 \centering
    \includegraphics[width = 8.5 cm]{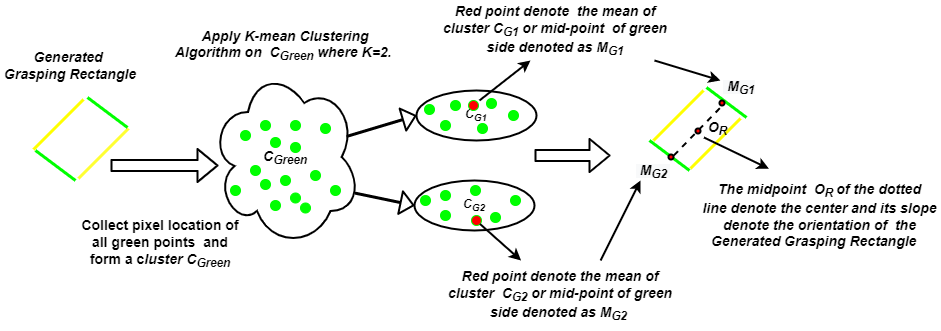}
    \caption{  Functioning of the Pose Extraction module}
    \label{fig:9}
\end{figure}

Fig. \ref{fig:9} explains the functioning of the pose extraction module. For finding the orientation of  a grasping  rectangle generated by Pix2Pix GAN, we have computed the slope of its yellow coloured side (other colours could also be used without loss of generality). The process followed are  firstly, the pixel locations are collected for all the green color (other colours could also be used without loss of generality) points of the generated grasping rectangle and then a cluster, named as $C_{Green}$,  is formed. Then,  K-Mean clustering algorithm has been applied on the cluster  $C_{Green}$  with the  k value taken as 2. As a result at the output two clusters  $C_{G1}$ and $C_{G2}$ are obtained that refer to the green sides of the generated grasping rectangle along with their mean value point. 

The mean point of the $C_{G1}$ and $C_{G2}$ cluster is denoted as $M_{G1}$ and $M_{G2}$ respectively, which also refer to the mid-point of the green sides of the generated grasping rectangle. So, the line joining  the mid-points $M_{G1}$ and $M_{G2}$ becomes parallel to the yellow sides of the generated grasping rectangle which is denoted by a dotted line as shown in Fig. \ref{fig:9}.
The slope of the dotted line refers to the orientation of the generated grasping rectangle and its mid point $O_R$ refers to the centroid of the generated grasping rectangle.
Algorithm \ref{algo:1} shows the functioning of our pose extraction module.


\begin{algorithm}
    \caption{Algorithm for Pose Extraction Module}
    \textbf{Input:}\\
    I - Image of an object tagged with the grasping rectangle generated by Pix2Pix GAN \\
    \textbf{Output:} \\
    Pose (position and orientation) of the generated grasping rectangle by Pix2Pix GAN\\
    \textbf{Procedure:}
    \label{algo:1}
    \begin{algorithmic}[1]
   
        \STATE {Collect location of all green pixel  from input image I and form a cluster $C_{Green}$.}
        
        \STATE {Apply K-Mean algorithm on  $C_{Green}$  where $k=2$.}
            
        \STATE {The output of K-Mean algorithm on $C_{Green}$ are two clusters i.e. $C_{G1}$ and $C_{G2}$  along with the mean point $M_1$ and $M_2$ respectively.}
        
        \STATE {Clusters $C_{G1}$ and $C_{G2}$ refer to green line of grasp rectangle, so mean point $M_1$ and $M_2$ refer to mid-point of green lines.}
        
        \STATE {The slope ($\theta$) of the line joining the mid point $M_1$ and $M_2$ of the green sides of the grasp rectangle is the  orientation of the optimal grasp rectangle which is calculated by using:}
        \begin{center}
            $\theta= \atantwo ( \triangle y, \triangle x )$
        \end{center}
        where
        \begin{center}
           $ \triangle y = [ M_1(y) - M_2(y)]$ \\ 
           $ \triangle x = [ M_1(x) - M_2(x)]$
        \end{center}
        
        \STATE {The mid point $O_R$  of the line joining the  mid point $M_1$ and $M_2$ of the green sides is the center of the optimal grasp rectangle which is computed by using:} 
        \begin{center}
            $O_{R_x} = [ M_1(x) + M_2(x)]/2$ \\
            $O_{R_y} = [ M_1(y) + M_2(y)] /2$
        \end{center}

\end{algorithmic}
\end{algorithm}


   \par
 After extracting the pose of the generated grasping rectangle, it is translated to an optimal position of the object  using pose translation module which has been discussed next.
 Firstly, the optimal position of the object is required to perform the translation of the generated grasping rectangle. We have logically assumed that the centroid  of any object as its optimal position, and is the primary position of interest for grasping,  which has been obtained by using the Image Moments of OpenCV \cite{bradski2000opencv}. The Image Moments is a weighted average of image pixel intensities, which  help in determining the specific property of image such as area, volume, centroid, radius etc.
(\ref{eq:10}) shows how the centroid of the object has been calculated.
 
\begin{equation}
\begin{split}
c_{x}= M_{10}/ M_{00} \\ 
 c_{y}= M_{01}/ M_{00}
\end{split}
\label{eq:10}
\end{equation}
where $c_x$ is the x-coordinate, $c_y$ is the y-coordinate of the centroid of the object  $O_O$ and M refers to the Moment.

 The following steps are performed for calculating the centroid of the object in an image using OpenCV:
 \begin{itemize}
     \item Convert the RGB image to gray-scale.
     \item Binarize the gray-scale image.
     \item Compute the Image Moments.
     \item Obtain the centroid using Moments. 
 \end{itemize}
 
\begin{figure}[!ht]
\centering
    \includegraphics[width = 8.5 cm]{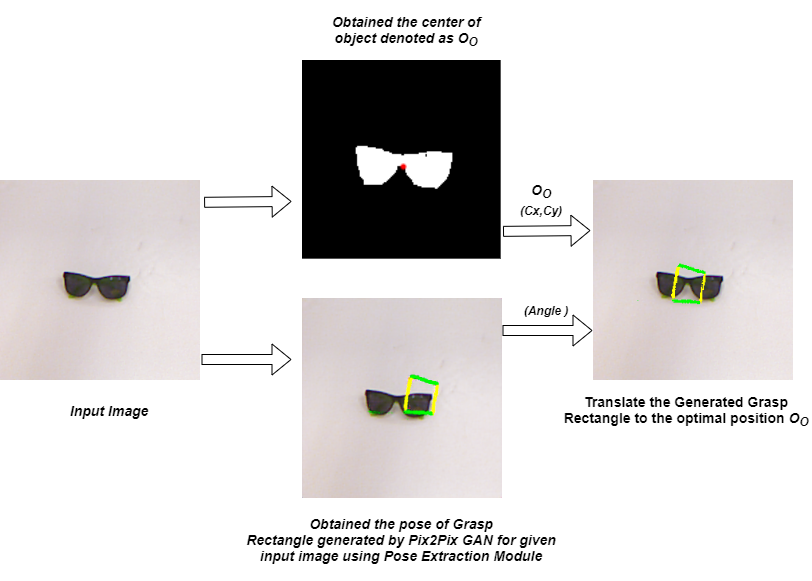}
    \caption{Functioning of the Pose Translation Module }
    \label{fig:10}
\end{figure}

After finding the centroid of the object, the generated grasping rectangle is translated  by adding the  distance vector between  the center of the generated grasping  rectangle $O_{R}$ and the center of the object $O_{O}$ to the sides of the generated grasping rectangles  as shown in Fig. \ref{fig:10}. Since computation of the  orientation of the grasping rectangle is purely based on the colors of the rectangle sides, to avoid ambiguity, during testing,  the objects having  green/yellow color have been avoided.
With the proposed methodology, we have created our indigenous dataset and its details have been discussed in subsequent section.  

\subsection{Procedure to create our indigenous dataset} 

\begin{figure}[!ht]
\centering
    \includegraphics[scale=.5]{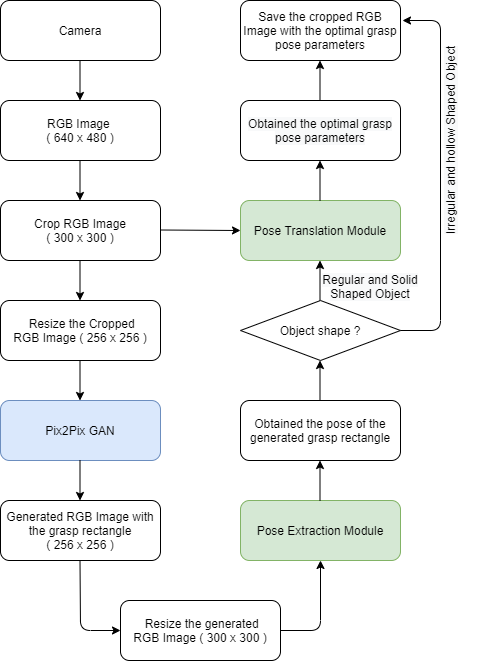}
    \caption{Procedure to create our dataset}
    \label{fig:11}
\end{figure}

In this section, we discussed about the whole process of creating robotic grasping dataset using our proposed approach which is its main purpose.
In Fig. \ref{fig:11}, the process of creating dataset is explained via a flow diagram. First, we have captured the images of different objects placed on the table at different locations with different orientation, using camera (Intel RealSense D435). Then we crop and resize the images and feed into the trained generative model i.e. Pix2Pix GAN. Our trained model generates the  grasping rectangle for the input images of the objects. Then with the help of pose extraction module, the pose (position and orientation) of the generated grasping rectangle is extracted. For the regular and solid shaped objects, we propose a translation module which translate the rectangle appropriately to the centroid  of the object making it an optimal grasping rectangle. However, for irregular and hollow shaped objects, it has been observed that such translation towards centroids are not needed and will not be appropriate as shown in Fig. \ref{fig:14} and thus for obtaining  the optimal grasping rectangles for such objects, the generator produced grasping rectangles are considered as it is for solving further grasp execution problems. After finding the optimal grasping rectangle for the object, a `slight refinement' has been used only to have a proper shaped rectangle with the fixed values of height and width $(h,w)$ for the rectangle, since the rectangle's height and width  play no role in passing the parameters $( x,y, \theta)$ of the rectangle to the robot controller for grasp execution.
Ultimately, the parameters of the refined optimal grasping rectangle are saved into a text file with the same naming convention of its corresponding image file, for the convenience of identifying correctly the object specific grasping rectangles.

Using the above mentioned methodology, till now we have generated a dataset of size approx 500 images and the process is continuing. Our aim, first, is to generate image dataset comparable to the widely available Cornell dataset and compare the performance between them. Thereafter, our proposed method can automatically produce large dataset, required for training an intelligent grasp learning architecture fulfilling the so called ``Robot for Vision" concept.
\section{Experiments with Anukul Robot  and Analyses of the results}

\begin{figure}[!ht]
\centering
    \includegraphics[scale=0.4]{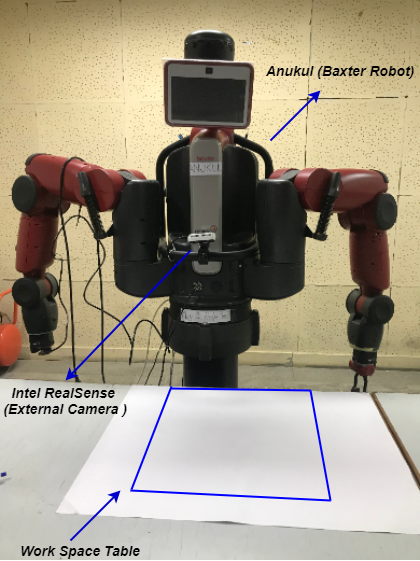}
    \caption{Experiment Setup}
    \label{fig:12}
\end{figure}

\subsection{Physical component}
Rigorous experiments have been performed for evaluating the performance of our proposed model in the real world environment. For grasp execution, a Research Robot (named as ``Anukul"/Baxter) developed by Rethink Robotics, has been used along with an external camera. Anukul is a two armed robot that has a 7-DOF with maximum reach of 104 cm. It is equipped with electrical and vacuum grippers. In our experiments, an electrical gripper is attached to the left arm of robot to perform the  grasping, although without any loss of generality the other hand could also be used. 

For an external camera, Intel RealSense D435 model is used, which consists of RGB sensor, depth sensor and infrared projector. It is mounted on the Anukul body at the height of 50 cm above the table and 45$^{\circ}$  facing  downward toward the table as shown in Fig. \ref{fig:12}. It gives the RGB-D image with a resolution of $640 \times 480$ pixels.

For each trail, an object is placed on the table, within the field of view of the mounted camera which has been shown as a blue square of size 40 cm $\times$ 40 cm. RGB image of the scene produce by external camera is first cropped to $300 \times  300$ and then resized to $256 \times 256$ for feeding into the trained model of Pix2Pix GAN for generating the grasping rectangle. After generation of the grasping rectangle, its position and orientation are computed using  pose extraction module and then with the help of pose translation module to obtain an optimal grasping rectangle,  the details of which have  already been discussed in the above section.

\subsection{Robotic Grasp Pose Estimation}
After finding an optimal grasping rectangle information in image space, it is further mapped with  the robot configuration space, so that robot can execute the grasp successfully. To perform the  mapping between the image configuration space and the robot  configuration space, we require two transformation matrices i.e. ${}^{R}T_{C}$ and ${}^{C}T_{I}$ as shown in (\ref{eq:2}). Firstly, image coordinates  are mapped to the camera coordinates  using (\ref{eq:11}), (\ref{eq:12}), (\ref{eq:13}). 

\begin{equation}
    x= [ (u - c_x ) / f_x ] * depth[u][v]
\label{eq:11}
\end{equation}

\begin{equation}
    y= [ (v - c_y ) / f_y ] * depth[u][v]
\label{eq:12}
\end{equation}

\begin{equation}
    z= depth[u][v]  
\label{eq:13}
\end{equation}

 which can be rewritten in the form of matrix as represented in the (\ref{eq:14}).
\begin{equation}
\begin{bmatrix}
u' \\
v' \\
w'
\end{bmatrix}
=
\begin{bmatrix}
f_x & 0 & c_x \\
0 & f_y & c_y \\
0 & 0 & 1
\end{bmatrix} 
\begin{bmatrix}
x\\y\\z
\end{bmatrix}
\label{eq:14}
\end{equation}
 
\noindent where, 
\begin{equation}
  u'= u * w' 
  \label{eq:15}
\end{equation} 

\begin{equation} 
  v'= v * w' 
  \label{eq:16}
\end{equation}

\begin{equation}
  w'= depth[u][v] 
  \label{eq:17}
\end{equation}

where $(u',v',w')$ represents the 3D point for 2D point $(u, v)$ in the image coordinates and mapped to point $(x,y,z)$  in the camera coordinates.

By observing the (\ref{eq:14}), we can conclude that the transformation matrix ${}^{C}T_{I}$ which consists of camera intrinsic parameters ( $f_x, f_y, c_x, c_y$)  can now be computed using (\ref{eq:18}).

\begin{equation}
{}^{C}T_{I}
=\begin{bmatrix}
f_x & 0 & c_x \\
0 & f_y & c_y \\
0 & 0 & 1
\end{bmatrix}^{-1} 
\label{eq:18}
\end{equation}

After finding the camera coordinates for the optimal grasp pose, it is mapped to the robotic grasp pose in the robot coordinate frame using transformation matrix ${}^{R}T_{C}$. For creating transformation matrix  ${}^{R}T_{C}$, 40 observations are taken on work-space table at the different locations, with the different orientations for ten different objects. Using transformation matrix  ${}^{R}T_{C}$, the position and orientation of any object with respect to the robot has been computed with the help of (\ref{eq:2}) which is denoted by $g_R$. 



\subsection{Grasp Execution in the Robot Space}
\begin{figure}[!ht]
\centering
    \includegraphics[scale=.4]{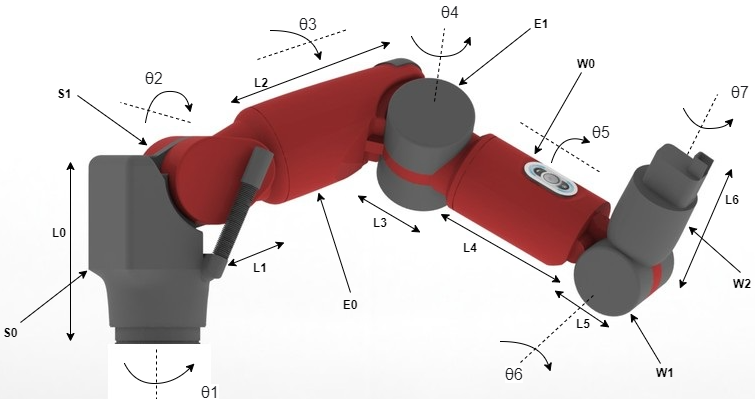}
    \caption{Anukul Robot Arm Joints}
    \label{fig:baxter}
\end{figure}

Once  robotic grasp pose $g_R$ is obtained, Anukul robot can execute the grasp, some details of which are discussed next.
For the determined  position and orientation, Anukul robot, which has 7-DOF as shown in Table \ref{tab:table1} and in Fig. \ref{fig:baxter}, first compute its joint angles by solving Inverse-Pose-Kinematics (IPK) using Denavit–Hartenberg (D-H) parameters as given in Table \ref{tab:table2} and \ref{tab:table3}.

\begin{table}

\begin{center}
\caption{Arm Configuration for Anukul Robot}
\scalebox{0.8}{
\begin{tabular}{|c|c|c|c|c|}
\hline 
 \textbf{Joint Motion} & \textbf{Joint Name} & \textbf{Joint Variable} &  \textbf{$\theta_i$ Min} & \textbf{$\theta_i$ Max}\\[5pt]
\hline 
  Shoulder Roll & $S_0$ & $\theta_1$ & $+51^{\circ}$ & $-141^{\circ}$ \\ [5pt]
\hline 
  Shoulder Pitch & $S_1$ & $\theta_2$ & $+60^{\circ}$ & $-123^{\circ}$ \\ [5pt]
 \hline  
  Elbow Roll  & $E_0$ & $\theta_3$ & $+173^{\circ}$ & $-173^{\circ}$ \\ [5pt]
\hline   
  Elbow Pitch & $E_1$ & $\theta_4$ & $+150^{\circ}$ & $-3^{\circ}$ \\ [5pt] 
\hline
  Wrist Roll  & $W_0$ & $\theta_5$ & $+175^{\circ}$ & $-175^{\circ}$ \\ [5pt]
\hline
  Wrist Pitch  & $W_1$ & $\theta_6$ & $+120^{\circ}$ & $-90^{\circ}$ \\ [5pt]
 \hline
  Wrist Roll & $W_2$ & $\theta_7$ & $+175^{\circ}$ & $-175 ^{\circ}$ \\ [5pt]
\hline
\end{tabular}
}
\label{tab:table1}
\end{center}

\end{table}

The 3-DOF are embedded into the wrist with a hope to obtain a closed-form kinematic solution as per Pieper's recommendation \cite{5649842}.

\begin{table}
\begin{center}
\caption{D-H Parameters for Anukul Robot Arm}
\scalebox{0.9}{
\begin{tabular}{|c|c|c|c|c|}
\hline 
\textbf{$i$}  & \textbf{$\alpha_{i-1}$ } & \textbf{$a_{i-1}$} (mm) & \textbf{$d_i$} (mm) &  \textbf{$\theta_{i}$} \\[5pt]
\hline 
 1 & $0^{\circ}$ & $0$ & $0$ & $\theta_1$  \\ [5pt]
\hline 
2 & $-90^{\circ}$ & $L_1$  & 0 & $\theta_{2} + 90^{\circ} $  \\ [5pt]
\hline 
 3 & $+90^{\circ}$ & 0 & $L_2$ & $\theta_3$  \\ [5pt]
\hline 
4 & $-90^{\circ}$ & $L_3$  & 0 & $\theta_4$  \\ [5pt]
\hline 
5 & $+90^{\circ}$ & 0 & $L_4$  & $\theta_5$  \\ [5pt]
\hline 
6 & $-90^{\circ}$ & $L_5$  & 0 & $\theta_6$  \\ [5pt]
\hline 
7 & $+90^{\circ}$ & 0 & 0 & $\theta_7$  \\ [5pt]
\hline 
\end{tabular}}
\label{tab:table2}
\end{center}
\end{table}

\begin{table}
\begin{center}
\caption{Anukul Robot Arm Link and Offset-Lengths}
\scalebox{0.9}{
\begin{tabular}{|c|c|}
\hline 
\textbf{Length}  & \textbf{Value (mm)} \\[5pt]
\hline 
  $L_0$ & 270.35 \\ [5pt]
\hline 
 $L_1$ & 69.00   \\ [5pt]
\hline 
 $L_2$ & 364.35  \\ [5pt]
\hline 
 $L_3$ & 69.00  \\ [5pt]
\hline 
$L_4$ & 374.29  \\ [5pt]
\hline 
$L_5$ & 10.00  \\ [5pt]
\hline 
$L_6$ & 368.30  \\ [5pt]
\hline 

\end{tabular}}
\label{tab:table3}
\end{center}
\end{table}

\subsubsection{Numerical Inverse Pose Solution for 7-DOF Anukul Robot Left Arm}
\emph{Given:} D-H parameters and the required end-effector position and orientation,
    
\begin{equation}
\Big[ {}^{0}T_{7} \Big]
=\begin{bmatrix}
r_{11} & r_{12} & r_{13} &   {}^{0}x_{7} \\
r_{21} & r_{22} & r_{23} &   {}^{0}y_{7} \\
r_{31} & r_{32} & r_{33} &   {}^{0}z_{7} \\
0 & 0 & 0 &   1
\end{bmatrix}
\label{eq:19}
\end{equation}

\emph{Calculate:} the joint angles ($\theta_1$, $\theta_2$, $\theta_3$, $\theta_4$, $\theta_5$, $\theta_6$, $\theta_7$) to reach the given end-effector pose.\\
We use numerical iterative Newton-Raphson (N-R) method to solve this IPK problem, the details of which can be obtained in \cite{phd_thesis}.

The vectors for N-R method implemented are:
    
\begin{equation}
 [ G ( \Vec{\theta} )]
=\begin{bmatrix}
    g_1 ( \Vec{\theta_1} ) \\
    g_2 ( \Vec{\theta_2} ) \\
    g_3 ( \Vec{\theta_3} ) \\
    g_4 ( \Vec{\theta_4} ) \\
    g_5 ( \Vec{\theta_5} ) \\
    g_6 ( \Vec{\theta_6} ) \\
    g_7 ( \Vec{\theta_7} ) 

\end{bmatrix} 
\label{eq:20}
\end{equation}

\begin{equation}
 [ \Vec{\theta} ]
=\begin{bmatrix}
    \Vec{\theta_1}  \\
    \Vec{\theta_2}  \\
     \Vec{\theta_3}  \\
     \Vec{\theta_4}  \\
     \Vec{\theta_5}  \\
     \Vec{\theta_6}  \\
     \Vec{\theta_7} 
\end{bmatrix} 
\label{eq:21}
\end{equation}

The N-R method involves numerical iterations to solve a coupled sets  of $m$ non-linear equations with $n$ unknowns.
\par 
Performing a Taylor Series expansion of $[G ( \Vec{\theta})]$ about $[\Vec{\theta}],$ we get:
\begin{equation}
\centering
\begin{split}
    g_{i} \Big( [\Vec{\theta}] + [\delta \Vec{\theta}] \Big) = g_{i} ([\Vec{\theta}]) + 
     \sum_{j=1}^{n}  (\delta g_i / \delta \theta_j)  * \delta \theta_j
    + O ( [ \delta \Vec{\theta}^2])
    \end{split}
    \label{eq:22}
\end{equation}

where $i= 1,2, ... m$. \\

Introducing   $ [J_{NR}] =  [J_{NR}(\theta)] = [\delta g_i / \delta \theta_j]$ as Newton-Raphson Jacobian  matrix and putting higher order term to zero i.e.
$O ( [ \delta (\Vec{\theta})^2]) \rightarrow 0$, we get:

\begin{equation}
\centering
\begin{split}
    g_{i} \Big( [\Vec{\theta}] + [\delta \Vec{\theta}] \Big) = g_{i} ([\Vec{\theta}]) +
     \sum_{j=1}^{n}  (\delta g_i / \delta \theta_j)  * \delta \theta_j \\
     = g_{i} ([\Vec{\theta}]) + [J_{NR}] [\delta \theta ] 
      \end{split}
      \label{eq:23}
\end{equation}
where $i= 1,2, ... m$.

To determine the compensation factor $\delta \Vec{\theta}$ at each iterative  solution step we need to solve the following equation:
\begin{equation}
    g_{i} ([\Vec{\theta}]) + [J_{NR}] [\delta \theta ] = 0
    \label{eq:24}
\end{equation}

 However, since Anukul robot has 7-DOF and hence it is a kinematically-redundant robot and the Newton-Rapshon Jacobian matrix is a non square one ($6 \times 7$), these equations are under-constrained as there are infinite solutions to the $\delta \Vec{\theta}$ vector at each iteration step.
 \\
 One possible solution is :
 \begin{equation}
    [J_{NR}]* [\delta \theta ] = -  g_{i} ([\Vec{\theta}])
    \label{eq:25}
\end{equation}

 \begin{equation}
   [\delta \theta ] = - [J_{NR}]^{-1} * g_{i} ([\Vec{\theta}])
   \label{eq:26}
\end{equation}

Using Moore-Penrose method for finding pseudo-inverse, we find the pseudo-inverse of Newton Rapshon Jacobian Matrix $[J_{NR}]^*$ :
\begin{equation}
   [J_{NR}]^* = [J_{NR}]^T \Big( [J_{NR}] [J_{NR}]^T \Big) ^{-1} 
   \label{eq:27}
\end{equation}
Note that  $[J_{NR}]^*$ is of size ($7 \times6$) which is pseudo-inverse of $[J_{NR}]$ having size ($6 \times 7$).

As it is well known, that the convergence success of N-R method clearly depends on initial guess. In the present situation the existing known pose configuration of Anukul robot  makes as an excellent initial guess which has been used for running Algorithm  \ref{algo:2}.


\begin{algorithm}
    \caption{Newton-Rapshon Algorithm for Inverse-Pose Solution }
    \textbf{Procedure:}
    \label{algo:2}
    \begin{algorithmic}[1]
   
        \STATE {Establish the function and variable to solve for: $[G(\Vec{\theta})]=0$.}
        
        \STATE {Make current robot configuration as initial guess to the solution: $[\Vec{\theta}]$.}
            
        \STATE {Solve $[J]([\Vec{\theta}]){\delta \theta_k}= [G(\Vec{\theta_k})]$ for ${\delta x_k}$, where k is the iteration counter.}
        
        \STATE {Update the current best guess for the solution: $(\theta_{k+1}) = (\theta_k)+(\delta \theta_k)$.}
        
        \STATE {Iterate until $|(\delta \theta_k)| < \epsilon$, where we use the Euclidean norm and $\epsilon$ is a small, user-defined scalar solution tolerance. Also, halt the iteration if the number of steps becomes too high (which mean the solution is diverging). }

\end{algorithmic}
\end{algorithm}


\subsubsection{Resolve-Rate control for Kinematically-Redundant Anukul robot}
\par

After getting joint solutions ($\theta_1$, $\theta_2$, $\theta_3$, $\theta_4$, $\theta_5$, $\theta_6$, $\theta_7$) from numerical inverse kinematics solution, it is required to calculate $ [\dot{\theta}$] from the well known relation given in \cite{craig2009introduction}.
\begin{equation}
    [\dot{\theta}] = [{}^{K}J(\theta)]^{-1} [{}^{K}\dot{X}]
    \label{eq:28}
\end{equation}
 Since we have used resolved-rate control strategy \cite{4081862}, for Anukul robot after defining required Cartesian velocities $[{}^{K}\dot{X}]$, the relative joint space velocity $[\dot{\theta}]$ is calculated using pseudo-inverse of the configuration dependent Jacobian matrix $[{}^{K}J(\theta)]^*$, by formulating the following optimization problem:
 \par Minimize the scalar cost function :
\begin{equation}
 f=(1/2 )* \Big( [\dot{\theta}]^T [\dot{\theta}] \Big)
 \label{eq:29}
 \end{equation}
    
    Subject to the rate equations:
\begin{equation}
    [\dot{X}]=[J][\dot{\theta}]
    \label{eq:30}
\end{equation}
which satisfies the primary task i.e. it provides the commanded Cartesian trajectories $[\dot{X}]$  as:
\begin{equation}
    [\dot{\theta_p}]=[{}^{K}J(\theta)]^{*}[{}^{K}\dot{X}]
    \label{eq:31}
\end{equation}
where $[{}^{K}J(\theta)]^{*}$ for Baxter robot is obtained from Moore-Penrose pseudo-inverse of the Jacobian Matrix.

\begin{equation}
    [J]^*=[J]^T \Big[ [J][J]^T \Big] ^{-1}
    \label{eq:32}
\end{equation}

\begin{figure}[!ht]
\centering
    \includegraphics[scale=.6]{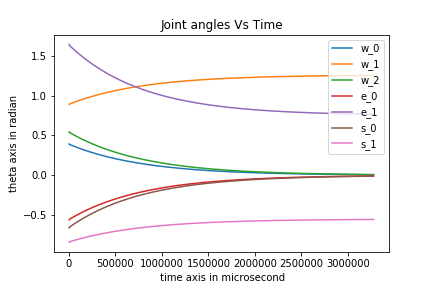}
    \caption{Joint  Angles Vs Time plot using Numerical Method}
    \label{fig:Plot}
\end{figure}

By closely looking the structure of $[J]^*$ from (\ref{eq:27}) and (\ref{eq:32}), we can infer that one very good reason of using numerical Inverse-Pose-kinematics solution for Anukul (instead of analytical solution) is that once we calculate $[J]^*$, it can be used both for numerical Inverse-Pose-kinematics solution as well as for Resolve-Rate control, there by saving computations.
Here, we are providing some grasp execution representative sample solutions in Table \ref{tab:theta} and Table \ref{tab:theta_dot} and in Fig. \ref{fig:Plot}.

\begin{table*}
\caption{Using Numerical method to compute Joint Angle $\theta_i$ (rad)}
\begin{center}
\begin{tabular}{|c|c|c|c|c|c|c|c|c|}
\hline 
\textbf{S.No.} & \textbf{$\theta_1$} & \textbf{$\theta_2$ } & \textbf{$\theta_3$ } & \textbf{$\theta_4$} & \textbf{$\theta_5$ } & \textbf{$\theta_6$ } & \textbf{$\theta_7$ } &  \textbf{t } (microsecond) \\ [5pt]
\hline 
1 &  0.39271744 & 0.88789328   & 0.54495699 & -0.56048391 & 1.64158602 & -0.66008341 &
 -0.83888404  & 4994071.0  \\ [5pt]
 \hline      
2 & 0.3878124 &  0.89253838 & 0.53815047 & -0.55348346 & 1.63044861 & -0.65183897 &
 -0.83527477  & 4994409.0  \\ [5pt]       
\hline  
$\dots$ & $\dots$ & $\dots$ & $\dots$ & $\dots$ & $\dots$ & $\dots$ & $\dots$  & $\dots$ \\ [5pt]
\hline 
100 & 0.11315895 & 1.15263627 & 0.15702577 & -0.16149975 & 1.00681916 & -0.1901987 &
 -0.63317734 & 5974718.0 \\ [5pt]
\hline 
101 & 0.11174559 & 1.15397472 & 0.15506452 & -0.15948262 & 1.00360999 & -0.18782311 &
 -0.63213736 & 5984744.0 \\ [5pt]       
\hline  
$\dots$ & $\dots$ & $\dots$ & $\dots$ & $\dots$ & $\dots$ & $\dots$ & $\dots$ & $\dots$ \\ [5pt]
\hline
328 & 0.00644394 &  1.2536958  &  0.00894197 & -0.00919675 &  0.76451154 & -0.01083104 & -0.55465355  & 8254702.0 \\ [5pt]
\hline
329 &  0.00636345 &  1.25377202 &  0.00883029 & -0.00908188 &  0.76432879 &  -0.01069576 & -0.55459433  & 8264700.0 \\ [5pt]
\hline
\end{tabular}
\end{center}
\label{tab:theta}
\end{table*}

\begin{table*}
\begin{center}
\caption{Using Numerical method to solve Resolve-Rate $\dot{\theta_i}$ (rad/microsecond) }
\scalebox{0.8}{
\begin{tabular}{|c|c|c|c|c|c|c|c|c|}
\hline 
\textbf{S. No.} & \textbf{$\dot\theta_1$ } & \textbf{$\dot\theta_2$} & \textbf{$\dot\theta_3$} & \textbf{$\dot\theta_4$} & \textbf{$\dot\theta_5$} & \textbf{$\dot\theta_6$} & \textbf{$\dot\theta_7$} &  \textbf{t} (microsecond)\\
\hline 
1 & -1.45119551e-05 &  1.37428784e-05 & -2.01376117e-05 & 2.07113727e-05 & -3.29509155e-05 &  2.43918397e-05 & 1.06782888e-05 & 4994409.0\\ [5pt]
\hline 
$\dots$ & $\dots$ & $\dots$ & $\dots$ & $\dots$ & $\dots$ & $\dots$ & $\dots$  & $\dots$ \\ [5pt]
\hline   
100 & -1.40969008e-07 &  1.33498203e-07 & -1.95616588e-07 &  2.01190098e-07 & -3.20084912e-07 &  2.36942122e-07 & 1.03728806e-07 & 5984744.0 \\ [5pt] 
\hline 
$\dots$ & $\dots$ & $\dots$ & $\dots$ & $\dots$ & $\dots$ & $\dots$ & $\dots$  & $\dots$ \\ [5pt]
\hline   
328 & -8.05008955e-09 &  7.62346638e-09 & -1.11707607e-08 &  1.14890381e-08 & -1.82785723e-08 &  1.35306712e-08 &  5.92347347e-09 & 8264700.0 \\ [5pt]
\hline
\end{tabular}}
\label{tab:theta_dot}
\end{center}
\end{table*}

\subsection{Experimental results and  Analyses}

Here, we have presented some analyses on results obtained during real time experiments involving Anukul robot, following the techniques described in section 5.2 and 5.3. We have tested the  performance of our model on Cornell Grasping Dataset as well as on our test-objects as  shown in Fig. \ref{fig:15}. 

\subsubsection{Performance evaluation on Cornell Grasping Dataset} 
To evaluate the performance of  our model we have classified the Cornell Grasping Dataset into two categories: regular and solid shaped objects and irregular and hollow shaped objects. 

\begin{figure*}
\centering
    \includegraphics[scale=0.55]{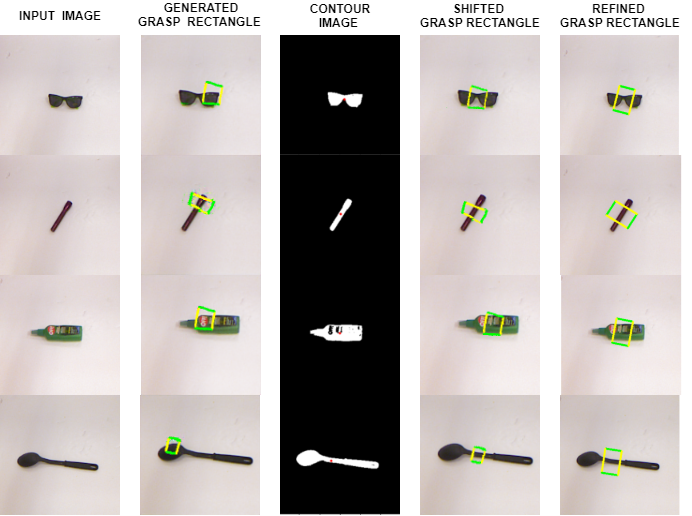}
    \caption{For regular and solid shaped objects of  Cornell Grasping Dataset, the generated grasping rectangle is translated \textbf{correctly} by our developed module to obtained the optimal grasping rectangle. }
    \label{fig:13}
\end{figure*}

\begin{figure*}
\centering
    \includegraphics[scale=0.55]{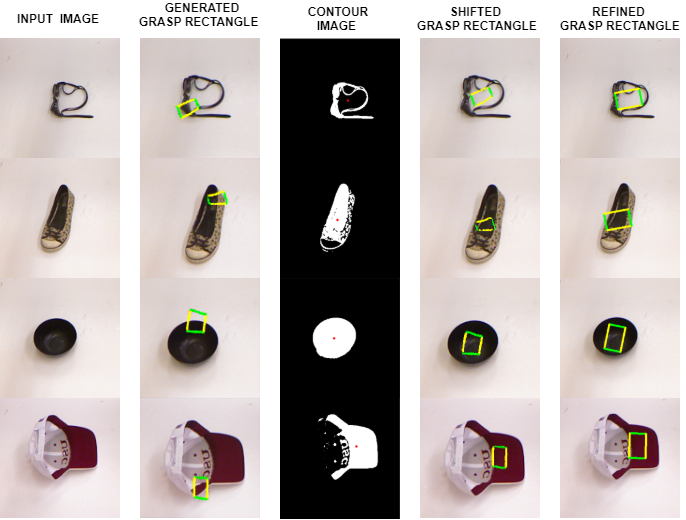}
    \caption{For irregular and hollow shaped objects of Cornell Grasping Dataset, the generated grasping rectangle is translated \textbf{incorrectly} by our developed module to obtained the optimal grasping }
    \label{fig:14}
\end{figure*}

Fig. \ref{fig:13} and Fig. \ref{fig:14} show the results obtained by our proposed approach for regular and solid shaped and irregular and hollow shaped objects  of the Cornell Grasping Dataset respectively. In both the figures, the first column shows the different types of input images taken by our trained model i.e. Pix2Pix GAN. The second column shows the output image of Pix2Pix GAN with the generated grasping rectangle. The third column  consists of the contour images with the  centroid of the object for the given input images. As it can be seen from both the pose extraction and pose translation modules, the generated grasping rectangle is translated to the centroid of the object to obtained the optimal grasping rectangle which has been  shown in the fourth column. Finally in the fifth column, the optimal grasping rectangles have been  refined and fine tuned  to  the proper shaped rectangles to make it suitable for successful robotic grasp execution.

For object grasping, we have hypothesized that like humans, robots also need to grasp solid and regular shaped objects towards their centroids. Centroids together with grasping rectangles would unambiguously help robots to grasp objects stably with minimum  grasping force (as grasping near the center of mass/centroid for regular shaped objects, from a mechanics point of view, requires minimum grasping force which is commensurate also with our daily experience of grasping objects). Our pose translation module is a step towards ensuring this. However, for other objects including hollow and irregular shaped objects, the output of the proposed architecture can be used straightway as an input to the robot grasping module of the robot hardware as  is evident in  Fig. \ref{fig:14}. Thus for the objects having  irregular and hollow shapes, the output need not pass through our pose translation module and as such there is no failure case as long as objects are having regular and irregular shapes. Only failure cases may occur when the objects are having irregular shapes with substantial masses which is unlikely for most of the industrial robots including cobots.

\subsubsection{Performance evaluation on testing object set} 

\begin{figure}[!ht]
\centering
    \includegraphics[width = 7.5 cm]{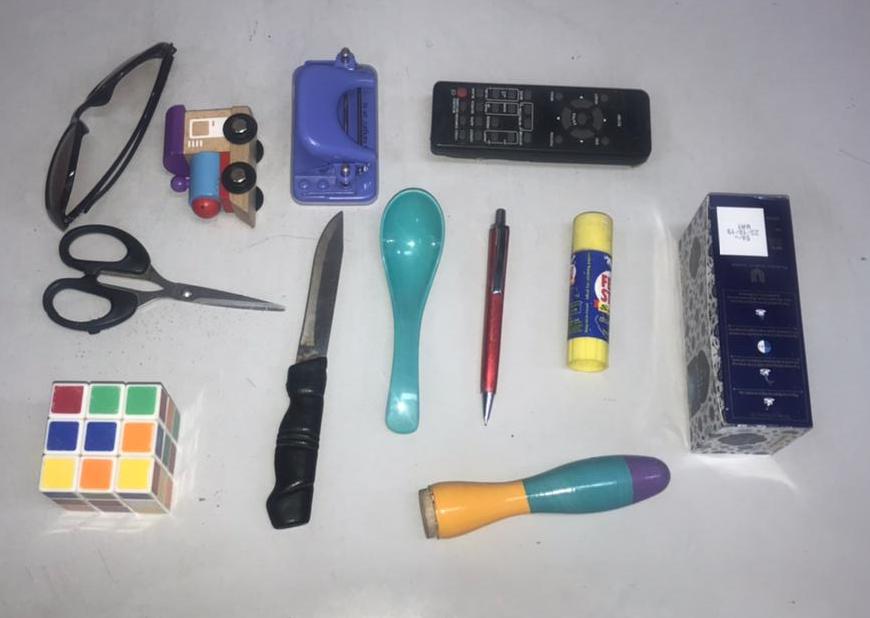}
    \caption{Testing object set for evaluating the performance our model.}
    \label{fig:15}
\end{figure}
\begin{figure}
\centering
    \includegraphics[scale=0.45]{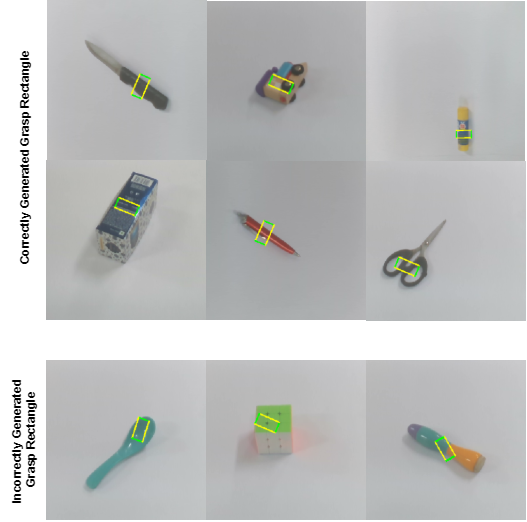}
    \caption{Grasping rectangle generated by our model for testing object set.}
    \label{fig:16}
\end{figure}

Here, we are testing the performance of our proposed approach by preforming the robotic experiments on the given testing object set shown in Fig. \ref{fig:15} in the real-world environment. Fig. \ref{fig:16} shows results obtained by our proposed approach for testing object set. Top rows in the figure shows the example of the correct optimal grasping rectangles and the bottom row depicts the failed examples.

By carefully analysing the results, we can say that the grasping rectangles generated for
regular and solid shaped objects by Pix2Pix GAN generative model, using our proposed methodology, is quite successful. However, in the case of irregular and hollow shaped objects the translation part of our model is not so effective and it should be bypassed for such objects, allowing our generative model created rectangles to remain as it is.

We have also computed the accuracy of our model on the  Cornell Grasping Dataset using the rectangle metric for classifying the grasping rectangle as discussed in \cite{jiang2011efficient}. 
Table \ref{tab:acc} shows the comparison between our results with the previous research using their self-reported scores for predicting the grasping rectangle accuracy on Cornell Grasping Dataset. It can be seen, that the performance of our generative model is very much comparable to the state-of-the-art discriminative models in terms of accuracy.

\begin{table}
\begin{center}
\caption{Grasp Rectangle Prediction Accuracy on the Cornell Grasp Dataset}
\begin{tabular}{|c|c|c|}
\hline 
\textbf{Author} & \textbf{Algorithm} & \textbf{Accuracy} \\
\hline 
- & Chance & 6.7 \%  \\ [5pt]
\hline 
Jiang et al.\cite{jiang2011efficient} & Fast-Search & 60.5\%  \\ [5pt]
 \hline  
Lenz et al. \cite{lenz2015deep} & SAE, stuct, reg, two-stage  & 75.6\% \\ [5pt]
\hline   
Redmon et al. \cite{redmon2015real} & AlexNet, Multi-Grasp & 88.0 \% \\ [5pt] 
\hline
Ours & Pix2Pix GAN & \textbf{87.79}\% \\ [5pt]
\hline
\end{tabular}
\label{tab:acc}
\end{center}
\end{table}

\section{Conclusion and Future work}

 We have presented a new approach in this paper for generating the grasping pose/rectangle from RGB images. By using image-to-image translation skill of Pix2Pix GAN, the grasping pose is generated directly for the input image of an object, instead of generating and selecting the best one from the candidate rectangles. 
 We have developed two modules to obtain an optimal grasping rectangle so that our model can enable successful robot grasp for both regular as well as irregular objects. With the help of the first module, the pose (position and orientation) of the generated grasping rectangle is extracted from the output of Pix2Pix GAN, and then the extracted grasp pose is translated to the centroid of the object, since here we hypothesize that like the  human way of grasping, for regular shaped objects centroids are the best locations/positions for stable grasping. For other irregular shaped objects, we allow the generated  grasping rectangles as it is for grasp execution.
 The performance of the proposed method is very much comparable to the state-of-art methods. To perform the grasp execution by a Anukul like redundant robot, we propose to use Numerical Inverse-Pose solution together with Resolve-Rate control so as to reduce computation time due to the sharing of data from the same configuration dependant Jacobian matrix. The experiments performed on real robot show that the proposed model could generate successful grasps for regular as well as irregular shaped objects including novel objects in the real time.

 For the training of our model only RGB images are used. To  improve the performance of the generative model in future, the depth images can also be incorporated during the training process. The present work is a significant step towards robot learning by  generating data sets, required for the huge number of parameters training in a deep learning architecture, automatically, reducing the drudgery of creating dataset manually, there by, making  the ``Robot for Vision" concept a reality in near future.

\section*{Acknowledgements}
The present research is partially funded by the I-Hub foundation for Cobotics (Technology Innovation Hub of IIT-Delhi setup  by the Department of Science and Technology, Govt. of India).

\section*{Conflict of interest}
The authors declare that they have no conflict of interest.


\bibliographystyle{spmpsci}      
\bibliography{bib}   

\end{document}